\def\year{2021}\relax
\documentclass[letterpaper]{article}
\usepackage{aaai21}
\usepackage{times}
\usepackage{helvet}
\usepackage{courier}

\usepackage{natbib}
\usepackage{caption}

\usepackage[hyphens]{url}
\usepackage{graphicx}
\graphicspath{ {figures/} }

\usepackage{amsmath}
\usepackage{subcaption}
\usepackage{booktabs}

\usepackage{textcomp}
\usepackage{scalerel}

\def\beamingEmoji{\scalerel*{\includegraphics{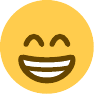}}{\textrm{\textbigcircle}}}
\def\blowingKiss{\scalerel*{\includegraphics{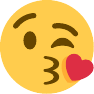}}{\textrm{\textbigcircle}}}
\def\eyes{\scalerel*{\includegraphics{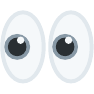}}{\textrm{\textbigcircle}}}
\def\rollingEyes{\scalerel*{\includegraphics{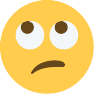}}{\textrm{\textbigcircle}}}
\def\fire{\scalerel*{\includegraphics{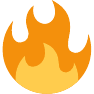}}{\textrm{\textbigcircle}}}
\def\foldedHands{\scalerel*{\includegraphics{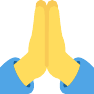}}{\textrm{\textbigcircle}}}
\def\greenHeart{\scalerel*{\includegraphics{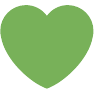}}{\textrm{\textbigcircle}}}
\def\heartEyes{\scalerel*{\includegraphics{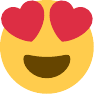}}{\textrm{\textbigcircle}}}
\def\hundredPoints{\scalerel*{\includegraphics{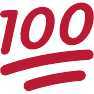}}{\textrm{\textbigcircle}}}
\def\loudlyCrying{\scalerel*{\includegraphics{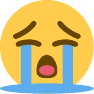}}{\textrm{\textbigcircle}}}
\def\pleadingFace{\scalerel*{\includegraphics{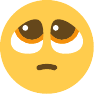}}{\textrm{\textbigcircle}}}
\def\purpleHeart{\scalerel*{\includegraphics{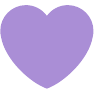}}{\textrm{\textbigcircle}}}
\def\redHeart{\scalerel*{\includegraphics{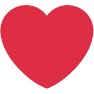}}{\textrm{\textbigcircle}}}
\def\rofl{\scalerel*{\includegraphics{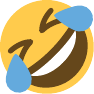}}{\textrm{\textbigcircle}}}
\def\skull{\scalerel*{\includegraphics{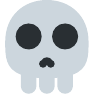}}{\textrm{\textbigcircle}}}
\def\smilingEyes{\scalerel*{\includegraphics{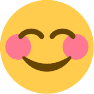}}{\textrm{\textbigcircle}}}
\def\smilingFaceHearts{\scalerel*{\includegraphics{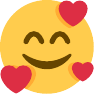}}{\textrm{\textbigcircle}}}
\def\sunglasses{\scalerel*{\includegraphics{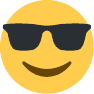}}{\textrm{\textbigcircle}}}
\def\tearsJoy{\scalerel*{\includegraphics{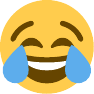}}{\textrm{\textbigcircle}}}
\def\thinkingFace{\scalerel*{\includegraphics{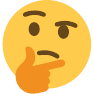}}{\textrm{\textbigcircle}}}
\def\thumbsUp{\scalerel*{\includegraphics{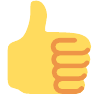}}{\textrm{\textbigcircle}}}
\def\sparkles{\scalerel*{\includegraphics{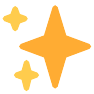}}{\textrm{\textbigcircle}}}
\def\winkingFace{\scalerel*{\includegraphics{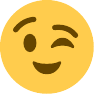}}{\textrm{\textbigcircle}}}
\def\wearyFace{\scalerel*{\includegraphics{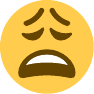}}{\textrm{\textbigcircle}}}
\def\twoHearts{\scalerel*{\includegraphics{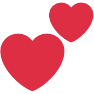}}{\textrm{\textbigcircle}}}
\def\monocleEmoji{\scalerel*{\includegraphics{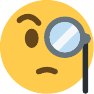}}{\textrm{\textbigcircle}}}
\def\droolingEmoji{\scalerel*{\includegraphics{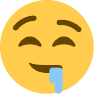}}{\textrm{\textbigcircle}}}
\def\crazyEyes{\scalerel*{\includegraphics{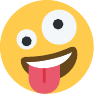}}{\textrm{\textbigcircle}}}

\title{The Shadowy Lives of Emojis:\protect\\An Analysis of a Hacktivist Collective’s Use of Emojis on Twitter}
\author{Keenan Jones, Jason R.~C.~Nurse, Shujun Li\\}
\affiliations{
Institute of Cyber Security for Society (iCSS) \& School of Computing\\ 
University of Kent, UK \\
\{ksj5, j.r.c.nurse, s.j.li\}@kent.ac.uk}

\begin{document} 

\maketitle 

\begin{abstract} 
Emojis have established themselves as a popular means of communication in online messaging. Despite the apparent ubiquity in these image-based tokens, however, interpretation and ambiguity may allow for unique uses of emojis to appear. In this paper, we present the first examination of emoji usage by hacktivist groups via a study of the Anonymous collective on Twitter. This research aims to identify whether Anonymous affiliates have evolved their own approach to using emojis. To do this, we compare a large dataset of Anonymous tweets to a baseline tweet dataset from randomly sampled Twitter users using computational and qualitative analysis to compare their emoji usage. We utilise Word2Vec language models to examine the semantic relationships between emojis, identifying clear distinctions in the emoji-emoji relationships of Anonymous users. We then explore how emojis are used as a means of conveying emotions, finding that despite little commonality in emoji-emoji semantic ties, Anonymous emoji usage displays similar patterns of emotional purpose to the emojis of baseline Twitter users. Finally, we explore the textual context in which these emojis occur, finding that although similarities exist between the emoji usage of our Anonymous and baseline Twitter datasets, Anonymous users appear to have adopted more specific interpretations of certain emojis. This includes the use of emojis as a means of expressing adoration and infatuation towards notable Anonymous affiliates. These findings indicate that emojis appear to retain a considerable degree of similarity within Anonymous accounts as compared to more typical Twitter users. However, their are signs that emoji usage in Anonymous accounts has evolved somewhat, gaining additional group-specific associations that reveal new insights into the behaviours of this unusual collective.
\end{abstract}

\section{Introduction}

The hacktivist collective Anonymous is an unusual one. Contrary to typical social groups, affiliates of Anonymous eschew notions of social hierarchy, membership, and set interests~\cite{Uitermark2017}. Instead, the group declares itself leaderless, an entity whose actions are dictated by the swarm-like movement of individual affiliates towards a given operation or `Op'~\cite{Olson2013}. Unlike most hacktivist groups, Anonymous maintains a clear public facing image, engaging with journalists and maintaining a high level of activity on social media sites, including Twitter~\cite{Beraldo2017}. Whilst some studies have focused on Anonymous' behaviours on social media, these studies have tended to take a higher level approach, examining how large-scale behaviours on Twitter from Anonymous affiliates relate to the overall philosophies of the group~\cite{Beraldo2017,Jones2020,McGovern2020}.

There also exists a number of studies focused on examining differences in emoji usage, but these typically focus on differences apparent in larger demographics, such as across different language speakers~\cite{barbieri2016,lu2016}. However, there is less research focused on the stability of emoji usage in online groups inhabiting a given social media platform. Given the surge in politically and socially relevant online groups in recent years, e.g., Anonymous, the Occupy movement, and Black Lives Matter, it is of great interest to examine whether the apparent `ubiquity' of emojis as a means of communication maintains itself in the often atypical behaviours of these groups~\cite{lu2016}.

To this, we present the first examination of emoji usage by a hacktivist collective, using a large network of Anonymous-affiliated Twitter accounts as a case study. In turn, we seek to answer the following question: \textbf{Are there any discernible differences in emoji usage on Twitter between Anonymous accounts and more `typical' users?} To do this, we:

\begin{itemize}
    \item Utilised Word2Vec models to compare how emojis are semantically related to each other. This found clear distinctions between emoji-emoji relations in Anonymous and non-Anonymous tweets.
    
    \item Applied sentiment analysis to identify and compare the emotional context in which emojis are used. Here we note that despite differences in emoji-emoji relations, and despite the range of sentiments that these emojis are used to convey, strong consistencies exist in the range of emotions being expressed by emojis in Anonymous and non-Anonymous tweets.
    
    \item Used qualitative evaluation of the most semantically relevant text tokens to label each emoji. In turn, we identified emojis which have received more `specified' usage by hacktivist Twitter accounts.
\end{itemize}

By taking these results together, we are able to present a clear picture of how accounts affiliated with one of the world's most prominent hacktivist groups utilise emoji, and how this usage compares to more `typical' Twitter users.

\section{Related Work}

Given the unusual nature of Anonymous and their public-facing online presence, considerable work has been done to gain further insights into the group. In \cite{Beraldo2017}, the authors analysed the evolution of a network of Twitter accounts broadcasting ``\#Anonymous''. Analysing this network's evolution over a period of three years, the authors identified consistently low stability in account usage of ``\#Anonymous''. This, in turn, fits with the group's claims of having an amorphous structure with no formal membership. In \cite{McGovern2020}, the authors continued this analysis of accounts linked to ``\#Anonymous'', studying how gender affected account posting behaviours. They found that male accounts showed a broad focus on group `Ops', whilst female accounts typically focused only on `Ops' related specifically to animal welfare.

Finally, \citet{Jones2020} focused on Twitter accounts specifically affiliated with Anonymous. Using a network of 20,000 Anonymous accounts, they conducted social network analysis to examine influence in the network. They found that the group showed signs of having a small set of highly influential accounts, a finding which contradicted the group's claims of having no set group of leaders.

Beyond Anonymous, a number of studies have focused on examining how emoji usage differs between groups. We detail a few notable works in this area. In \cite{lu2016}, the authors examined the usage of emojis in text messaging, using statistic-based analyses to examine emoji usage by nationality. Their work found that individual emojis followed a skewed distribution, with the most popular emojis accounting for the majority of emoji usage. Co-occurrence of emojis was also measured, finding indications of distinct patterns of emoji co-occurrence in certain countries.

In \cite{barbieri2016W2V}, the authors experimented with the utility of the popular Word2Vec (W2V) technique in modelling emoji usage. They tested a series of W2V models trained on Twitter data, examining the effects of data pre-processing and hyperparameter tuning on W2V's ability to model emoji use. They found that W2V is useful in this context, demonstrating the ability to learn the semantic relationships of emojis to other emojis and text items in a corpus. 

A similar study was conducted by \citet{reelfs2020}, in which the authors tested the ability of W2V in modelling semantic emoji associations on the online social network Jodel. Using a qualitative analysis of the semantic emojis and text neighbours of each emoji in the dataset, the authors identified good indications of the ability of W2V embeddings to capture insightful semantic relationships between emojis and text items in online communications.

In turn, \citet{barbieri2016} used W2V to study differences in Twitter emoji usage by users speaking different languages. By examining the intersection between most similar emojis to a given input emoji across several languages, the authors found indications of stability in the semantics of popular emojis.

Additionally, \citet{hagen2019} explored emoji differences in two narrower sets of Twitter users that were explicitly pro- or anti-white nationalism. Using frequency analysis, the authors found that there were distinct patterns in emoji usage present, with anti-white nationalism accounts using emojis such as ``Water Wave'' to represent the US democratic blue wave victories in 2018, and pro-white nationalism accounts using emojis such as ``Red X'' to indicate solidarity against shadow-banning.

\section{Contributions}

Our work builds on the findings in \cite{hagen2019} that emoji usage may have particular meaning distinct to a given group of Twitter accounts.

To this, we present the first examination of emoji usage by a large network of Anonymous-affiliated Twitter accounts, comparing it to a large random sampling of non-Anonymous Twitter users. Given the apparent ``new wave'' of hacktivism in recent months~\cite{reuters2021}, and Anonymous' apparent recent resurgence and unusual public-facing image ~\cite{Independant2020}, it is of particular interest to see if the group displays their own idiosyncratic patterns of emoji usage. This will provide new insights into the manner in which these affiliates utilise the Twitter platform, going beyond past studies which focused primarily on the broad structural patterns and interests of the group. Moreover, our study also provides unique insights into the notions of emojis as a ubiquitous and universal language~\cite{lu2016}, examining how well this holds within this unusual group.

Consequently, we present a multi-dimensional analysis of emoji usage, considering more typical notions including emoji frequency and emoji-emoji semantic relationships alongside analysis of emoji sentiment and context. Beyond Anonymous, this approach can also be leveraged to analyse emoji use by other noteworthy online groups (hacktivists or otherwise). Given the relevance of controversial online groups, such as QAnon~\cite{Papasavva2020}, over the past few years the flexibility of this approach could be of particular value to researchers interested in studying emoji usage within these groups. 

\section{Methodology}

In order to achieve this paper's aims, we used a series of computational measures to compare and contrast sentiment and semantic similarity in emoji usage between these two groups. From this, we aim to gain an understanding of the emotional purposes and typical contexts in which Anonymous accounts use emojis. In turn, our results provide insights into how the emoji presents itself within this atypical hacktivist group relative to more `typical' Twitter users.

\subsection{Data Collection}

In order to compare Anonymous' emoji usage on Twitter, we collected two datasets: one comprised of tweets from Anonymous accounts, and the other comprised of tweets from a random sample of non-Anonymous Twitter accounts. The random baseline dataset is then analysed alongside our Anonymous dataset to examine how Anonymous' usage of emojis compares to that of `typical' Twitter users.

As identifying a relevant set of Anonymous accounts is difficult, we utilised the pre-established method conducted by \citet{Jones2020} which drew on a set of five Anonymous seed accounts and utilised a combined approach of snowball sampling and machine learning classification to sample additional accounts. As one of these seed accounts has since been banned by Twitter, we added a further seed account: `@YourAnonCentral', which has been identified in recent news articles as being prominently linked with the group~\cite{AJC2020}.

A two-stage snowball sampling approach was then conducted, collecting the Anonymous followers and followees of these five seed accounts (Stage 1), and thereafter the Anonymous followers and followees of the newly identified set of Anonymous accounts (Stage 2). As the complete set of followers and followees is large (more than 10 million accounts at Stage 1), we utilised machine learning classification to identify Anonymous accounts at each stage. To do this, a dataset of accounts annotated as Anonymous and non-Anonymous was first needed to train our classifier.

From the complete set of followers and followees collected from the five seeds, we annotated accounts according to the established heuristic that an Anonymous account should have at least one Anonymous keyword in either its username \textbf{or} screen-name, \textbf{and} in its description, as well as having a profile or background image containing either a Guy Fawkes mask or a floating businessman (images commonly associated with Anonymous~\cite{Olson2013}). The Anonymous keywords were sourced from \cite{Jones2020}, and can be found in Table~\ref{table:keywords}.

\begin{table}[!tbh]
\centering
\begin{tabular}{ c c c c }
 \toprule
 anonymous & an0nym0u5 & anonymou5 & an0nymous \\
 anonym0us & anonym0u5 & an0nymou5 & an0nym0us \\ 
 anony & an0ny & anon & an0n\\ 
 legion & l3gion & legi0n & le3gi0n\\
 leg1on & l3g1on & leg10n & l3g10n\\
 \bottomrule
\end{tabular}
\caption{Anonymous keywords used.}
\label{table:keywords}
\end{table}

Firstly, we used keyword searches to filter accounts by the presence of at least one Anonymous keyword in either their username or screen-name. This yielded a set of 44,914 accounts. These accounts were then manually annotated in accordance with the above heuristic as being Anonymous or not. Given the filtering steps above, this annotation process was straightforward and simply required verification that the filtering steps had worked appropriately, and an examination of the profile and background images for either of the two Anonymous images selected at the definition stage. Initially, three annotators with substantial knowledge of the group annotated a subset of 200 accounts. Fleiss's Kappa was then used to calculate agreement, yielding a near-perfect score of 0.92. Given the high level of agreement, a single annotator from the three annotated the remaining accounts. This annotation process identified 11,349 Anonymous accounts and 33,565 non-Anonymous accounts. 

These accounts were then used to train a series of machine learning classifiers: SVM, random forest, and decision trees, using five-fold cross validation and the 62 features listed in \cite{Jones2020}. The results of this can be found in Table~\ref{table:ML_performances}. As random forest was the best performer, it was selected and trained on the complete set of 44,914 annotated accounts.

\begin{table}[!tbh]
\centering
\resizebox{.95\columnwidth}{!}{
\begin{tabular}{ c c c c }
\toprule
\textbf{Model} & \textbf{Precision} & \textbf{Recall} & \textbf{F1-Score} \\
\midrule
Random forest & \textbf{0.94} & \textbf{0.94} & \textbf{0.94} \\ 
Decision tree & 0.91 & 0.91 & 0.91\\ 
SVM (sigmoid kernel) & 0.67 & 0.74 & 0.67\\
\bottomrule
\end{tabular}
}
\caption{Performances of the three machine learning models.}
\label{table:ML_performances}
\end{table}

This trained model was then used to identify Anonymous accounts at each stage of the snowball sampling. This found 31,562 Anonymous accounts in the first stage and a further 11,013 accounts in the second, yielding a total of 42,575 Anonymous accounts. 

It should be acknowledged that the Anonymous definition used to annotate accounts is likely over prescriptive. Given their amorphous and inconsistent nature~\cite{Olson2013}, building an encompassing definition for Anonymous affiliates is impossible. Instead, we utilise the above strict definition which yields a set of Anonymous accounts that are (given the subject matter) relatively uncontroversial. Due to the strictness of the definition, however, it should be noted that the classification approach likely gives a large number of false-negatives, and thus the numbers found are not necessarily indicative of the `true' number of Anonymous accounts. With that being said, the set of accounts identified is sufficiently large that we can conduct analysis of the group with a good degree of confidence.

Having identified our set of Anonymous-affiliated accounts, Twitter's timeline API was used to retrieve the latest tweets from each Anonymous account (to a maximum of 3,200 tweets) as of 3rd December, 2020~\cite{TwitterAPI}. This provided a dataset of approximately 11 million tweets. We then filtered any tweets in the dataset not written in English to control for differences in emoji usage by speakers of different languages. We also filtered out retweets, as they likely do not reflect the emoji usage of the retweeting account. This resulted in a dataset of 4,709,758 tweets.

We then identified tweets in the dataset containing at least one emoji, finding 323,357 tweets. To account for any potential differences in language usage between accounts that use emojis and accounts that do not, we then extracted from our Anonymous dataset tweets from accounts that posted at least one tweet containing emojis. As the time-frame of post dates ranged from January 2007 to December 2020, we also limited our dataset to tweets posted in 2020 to remove any potential impacts caused by changes in emoji usage over time. This yielded a final Anonymous dataset of 980,587 tweets from 9,926 Anonymous accounts; this is the dataset that is the basis of this research. 

In order to examine any differences in emoji usage by Anonymous accounts, we compared them to a baseline of randomly sampled non-Anonymous Twitter users. Similar approaches have been used in the past to examine potential differences in language use amongst specific Twitter groups, including pro-ISIS accounts~\cite{torregrosa2020} and accounts from users suffering from PTSD~\cite{coppersmith2014}.

To provide the baseline dataset, we utilised Twitter's realtime sampling API to collect a set of randomly sampled tweets~\cite{TwitterAPI}. Each unique account collected from this realtime sampling was then extracted. In total, 12,576 accounts were sampled. Twitter's timeline API was then used to extract the latest tweets from each account. Just as with the Anonymous dataset, non-English tweets and retweets were filtered. Due to limitations on our timeline API usage, the extraction was conducted after the Anonymous data was collected, finishing in March 2021. Therefore, we filtered this dataset for tweets that had been posted from January 1, 2020 up to March 12, 2021. Although this dataset does not match exactly with the time-frame captured in the Anonymous dataset, the date ranges are similar enough that emoji usage is unlikely to have been effected.

To help confirm this, we examined the cosine similarity between the frequencies of the top 20 emojis in baseline tweets from 2021 and baseline tweets in 2020 (the period that overlaps with our Anonymous dataset). This identified a cosine similarity of 0.96, indicating a high degree of consistency in popular emoji choices. This strengthens our assumption, indicating that popular emoji usage is fairly stable in our dataset, and therefore that the slight difference in dataset time-frames is unlikely to have impacted our findings. Again, tweets containing emojis were identified, with 366,243 being found. Tweets in our baseline dataset from accounts with at least one emoji tweet were then extracted, resulting in 1,693,240 tweets from 9,180 accounts. 

This final set of accounts was then checked for the presence of any Anonymous accounts. To this, we first looked for any accounts in our complete Anonymous dataset that appeared in this dataset. We then utilised the Anonymous keyword search on the username, screen-name, and descriptions of these baseline accounts. Any accounts found in this search were then manually examined using our prescribed definition of an Anonymous account. Overall, this process flagged three accounts, none of which were found to be Anonymous-affiliated. Thus, the final dataset remained at 1,693,240 tweets from 9,180 accounts.

\subsection{Modelling Emoji Usage}

To model how emojis were used by both Anonymous Twitter users and randomly sampled Twitter users in both datasets, we opted to use the popular Word2Vec (W2V) approach~\cite{mikolov2013}. Whilst originally intended to model language usage, this approach has been found to be effective at also modelling emoji usage~\cite{barbieri2016W2V,reelfs2020}.

We constructed two W2V models, one for the Anonymous tweet dataset and the other for the random baseline tweet dataset. This would then allows us to learn the semantic relationships regarding the use of emojis present in each dataset, and thus compare the similarities and differences in emoji usage between the two Twitter user groups.

As past studies have shown that pre-processing approaches can be useful for improving the quality of emoji-focused W2V models~\cite{barbieri2016W2V}, we conducted a set of pre-processing measures on each tweet in each dataset prior to modelling. These included removing Twitter specific noise such as user tags, removing URLs, removing stop words, and expanding contractions. Lemmatisation was then used to assist each model in making connections between related terms.

Based on past studies~\cite{barbieri2016W2V} and our own experimentation, we settled on a vector size of 300 and a context window size of 6 as the optimum hyperparameter values for our models. We also tested varying values of the minimum count hyperparameter used to ignore tokens that occur infrequently. This was particularly necessary due to the noise inherent in Twitter data. We experimented with values between 3 and 15, choosing 10 as this was the value that produced the most interpretable results without discarding unnecessary data.

These models were then used to identify the most semantically similar emojis and text tokens to emojis used by Anonymous and baseline Twitter accounts using the cosine similarity between the most frequent emojis to extract their nearest emoji and text neighbours identified by each W2V model. The cosine similarity provides a measure of the semantic similarity between embeddings and thus allows for the identification of the most semantically similar tokens~\cite{barbieri2016W2V}.

We then utilised the Jaccard Index to provide a measure of the similarity of the two sets of most related emoji neighbours, for each emoji from the Anonymous and baseline W2V models. The Jaccard Index measures the number of shared members between two sets as a percentage. It is defined as follows:
\[ J(X,Y) = |X \cap Y|~/~|X \cup Y|, \]
where $X$ is the set of Anonymous emoji neighbours to a given emoji and $Y$ the set of baseline emoji neighbours to a given emoji. By doing this, we gain an understanding of the similarity between the semantic emoji neighbours in each dataset, and thus a sense of the similarities/differences in emoji usage.

\subsection{Sentiment Analysis of Emoji Usage}

In order to examine how similar emoji usage is between Anonymous and non-Anonymous accounts, it is not sufficient to measure emoji usage purely off of semantically similar tokens. Differences in semantic relations do not necessitate differences in the emotional context that a given emoji is used in. Given the noted importance of emojis as a means of communicating emotion~\cite{lu2016}, this aspect warrants consideration when examining emoji usage.

Therefore, we compare the emotion being expressed by tweets in each dataset containing emoji. To do this, we utilised the VADER sentiment analysis tool, a popular lexicon-based sentiment analysis model that is optimised for both tweet data and emoji analysis~\cite{Gilbert2014}. We then calculated the sentiment scores for each set of tweets from each dataset that contained at least one occurrence of a given emoji. This provided insights into the typical emotional context in which these emojis appear in each dataset, and allowed for comparisons of whether any similarities or differences are present in the emotional context of emoji tweets in Anonymous and baseline Twitter users.

Cohen's D was next utilised to measure differences in sentiment between emoji tweets from the two datasets~\cite{cohen2013}. Cohen's D measures the standardised difference between the means of two samples in terms of the number of standard deviations that the two samples differ by. Denoting the sizes of the two samples by $n_1$ and $n_2$ and their means by $\mu_1$ and $\mu_2$, Cohen's D is expressed as:
\[ d = \frac{\mu_1 - \mu_2}{s}, \text{ where } s = \sqrt{\frac{(n_1 - 1)s^2_1 +(n_2 - 1)s^2_2}{n_1 + n_2 - 2}}. \] 
In the above equations, $s$ is the pooled standard deviation of the two samples. By using this, we were able to gain an understanding of the degree of difference there is in the way in which emojis are used to convey emotions in Anonymous and non-Anonymous tweets.

\subsection{Ethics}

In order to ensure the ethical integrity of our study and to preserve the privacy of the users included, we ensured that all data collection was made in accordance with Twitter's API terms and conditions~\cite{TwitterDev}. We ensured to refrain from providing the names of any accounts included in this study, other than those that have already been included in published articles. Additionally, any direct quotes drawn from tweets have been published without attribution to protect the source account's privacy. We also ensure that any tweets quoted here come from accounts that do not contain any identifiable information in their Twitter bio. Moreover, only publicly available data was used in this study and any account that was deleted or suspended, or made protected or private was not included in the data collection process.

\section{Results and Discussion}

In this section we discuss the results of our study to compare emoji usage between Twitter accounts affiliated with the hacktivist collective Anonymous and Twitter accounts drawn from a random sample of all Twitter users.

\subsection{Emoji Frequencies}

In Table~\ref{table:emojiFrequencies}, we present the top 20 most popular emojis in both the Anonymous Twitter dataset and the baseline Twitter dataset\footnote{All emoji images are obtained from the open source Twemoji project (\url{https://twemoji.twitter.com/}), licensed under CC-BY 4.0.}. This table presents both the top 20 emojis in each dataset, as well as the raw counts of unique occurrences across tweets in each dataset and the percentage of total emoji usage that each emoji constitutes.

\begin{table}[ht!]
\centering
\begin{tabular}{c | c | c} 
 \toprule
 \textbf{Rank} & \textbf{Anonymous} & \textbf{Random Baseline} \\
 \midrule
 1 & 
 \tearsJoy~~32,679 (7.06\%) & \tearsJoy~~43,138 (8.31\%)\\
 2 & 
 \redHeart~~13,383 (2.89\%) & \loudlyCrying~~40,372 (7.78\%)\\
 3 & 
 \heartEyes~~12,905 (2.79\%)& \pleadingFace~~18,640 (3.59\%)\\
 4 & 
 \rofl~~9,289 (2.01\%) & \rofl~~16,817 (3.24\%)\\
 5 & 
 \loudlyCrying~~8,596 (1.87\%) & \redHeart~~15,607 (3.01\%)\\
 6 & 
 \thumbsUp~~7,645 (1.65\%) & \heartEyes~~10,587 (2.04\%)\\
 7 & 
 \smilingEyes~~7,544 (1.63\%) & \smilingFaceHearts~~9,276 (1.79\%)\\
 8 & 
 \greenHeart~~7,253 (1.57\%) & \fire~~8,535 (1.64\%)\\
 9 & 
 \fire~~6,323 (1.37\%) & \foldedHands~~6,383 (1.23\%)\\
 10 & 
 \sunglasses~~6,207 (1.34\%) & \sparkles~~6,144 (1.18\%)\\
 11 & 
 \winkingFace~~5,374 (1.16\%) & \wearyFace~~6,067 (1.17\%)\\
 12 & 
 \blowingKiss~~5,179 (1.12\%) & \smilingEyes~~5,659 (1.09\%)\\
 13 & 
 \eyes~~5,150 (1.11\%) & \thumbsUp~~5,437 (1.05\%)\\
 14 & 
 \foldedHands~~50,95 (1.10\%) & \purpleHeart~~5,352 (1.03\%)\\
 15 & 
 \twoHearts~~5,002 (1.08\%) & \skull~~5,178 (1.00\%)\\
 16 & 
 \thinkingFace~~4,821 (1.04\%) & \rollingEyes~~5,128 (0.99\%)\\
 17 & 
 \sparkles~~4,691 (1.01\%) & \greenHeart~~4,994 (0.96\%)\\
 18 & 
 \beamingEmoji~~4,342 (0.94\%) & \twoHearts~~4,964 (0.96\%)\\
 19 & 
 \rollingEyes~~4,265 (0.92\%) & \thinkingFace~~4,761 (0.92\%)\\
 20 & 
 \pleadingFace~~4,167 (0.90\%) & \hundredPoints~~4,674 (0.90\%)\\
 \bottomrule
\end{tabular}
\caption{Emoji frequencies for the top 20 emojis in the Anonymous and baseline Twitter datasets.}
\label{table:emojiFrequencies}
\end{table}

Interestingly, the majority of popular emojis across the two datasets are shared, with 15 of the 20 emojis occurring in the top 20 for both datasets and a cosine similarity of 0.83 being recorded for the two sets of frequencies. Moreover, they both follow similar patterns of usage, with the top emojis in both datasets receiving a disproportionate share of the total usage. This is particularly pronounced given that there were 887 and 1,226 distinct emojis used in the Anonymous dataset and the non-Anonymous baseline dataset, respectively. This finding presents some similarity to the results of \citeauthor{lu2016}'s study~(\citeyear{lu2016}) of emoji usage by smartphone users, in which the authors identified that emoji frequency followed a power law distribution similar to the one identified by us on Twitter. 

With that being said, despite similarities our results are less dramatic than those in \cite{lu2016}, with the difference between the top emojis and other emojis being less pronounced in both datasets. In \cite{lu2016}, the authors note that `119 out of the 1,281 emojis', or 9.28\% of emojis, constitute around 90\% of usage. In the random baseline dataset, 300 of the 1,226 (24.47\%) emojis constitute 90\% of total emoji usage, and the Anonymous dataset 350 of the 887 (39.46\%) distinct emojis constitute 90\% usage. 

Thus, although emoji usage appears to be biased towards a distinct subset of the total amount of emojis used, in both our Twitter datasets this is less pronounced. Additionally, we see here some separation between Anonymous emoji usage, compared to that of `typical' Twitter users. Anonymous users in total use a smaller range of emojis than those of our baseline set. However, within this smaller set Anonymous accounts seem to show less of a clear preference towards a subset of emojis, with the distribution of usage being more even than in the baseline data and the results identified in \cite{lu2016}. 

\subsection{Measuring Emoji to Emoji Similarity}
\label{secton:similarity}

Although our initial investigations of emoji frequencies between Anonymous and non-Anonymous accounts point to some minor differences in emoji usage, these results provide little insight into the manner in which these emojis are used. To further investigate this, we utilise the W2V models trained on each dataset to identify the most semantically similar emojis to each of the most popular emojis found in Table~\ref{table:emojiFrequencies}. Given the use of similar emoji-emoji analysis in examining differences in emoji usage between groups~\cite{barbieri2016,barbieri2016W2V}, this is a useful starting point for our comparison between Anonymous and baseline Twitter users.

In order to ensure that we were capturing the most semantically relevant neighbours, we used a cosine similarity threshold, only identifying emojis scoring above the threshold as being semantically related. We experimented with threshold values between 0.4 and 0.7, as these ensured that we were capturing neighbours with some degree of semantic relevance to each emoji. We report the results for thresholds 0.5 and 0.6 as these provided the most meaningful results.

Results of the Jaccard Index for the sets of semantically similar emoji neighbours (with a cosine threshold of 0.5) between our Anonymous and baseline data can be found in Table~\ref{table:EmojiJaccardSim}. If the Jaccard Index is high for two sets of emojis, this indicates that the semantic relationships of a given emoji between the two datasets is stable, suggesting similar usage. If the score of an emoji is low, this indicates that the usage of that emoji is not consistently defined across the two datasets.

\begin{table}[bht!]
\centering
\begin{tabular}{c | c | c} 
 \toprule
\multicolumn{3}{c}{Emoji (Jaccard Index)}\\
\midrule
\thinkingFace~(50.00\%) & \rofl~(9.09\%) & \tearsJoy~(2.78\%)\\
\loudlyCrying~(31.71\%) & \redHeart~(8.62\%) & \heartEyes~(2.65\%)\\
\rollingEyes~(27.27\%) & \twoHearts~(7.85\%) & \greenHeart~(1.85\%)\\
\wearyFace~(25.53\%) & \beamingEmoji~(7.41\%) & \sparkles~(0.95\%)\\
\pleadingFace~(20.59\%) & \sunglasses~(7.41\%) & \purpleHeart~(0.00\%)\\
\skull~(17.39\%) & \winkingFace~(6.82\%) & \fire~(0.0\%)\\
\smilingFaceHearts~(16.04\%) & \foldedHands~(6.78\%) & \eyes~(0.00\%)\\
\blowingKiss~(13.33\%) & \thumbsUp~(6.25\%)\\
\smilingEyes~(11.86\%) & \hundredPoints~(6.06\%)\\
\bottomrule
\end{tabular}
\caption{Jaccard Index between the nearest emoji neighbours identified by our W2V models, for the 25 most frequently used emojis in our datasets.}
\label{table:EmojiJaccardSim}
\end{table}

As we can see, the typical similarities in popular emoji usage between Anonymous and baseline users appears to be low, with the most similarly used emoji, `Thinking' (\thinkingFace), only achieving a similarity of 50\%. Additionally, we find that the majority of emojis receive a similarity score of less than 15\%. 

This result is somewhat surprising, given that both datasets seem to largely share the same set of frequently used emojis, and given that they both seem to use each emoji to a fairly similar degree. It thus seems that Anonymous Twitter users display fairly unique emoji-emoji definitions of these popular emojis, relative to that of Twitter users in general.

To ensure that these differences were not the result of the lower cosine similarity threshold (0.5), which may have introduced less relevant neighbours, we also experimented with a threshold of 0.6. With this higher threshold, we actually found that similarities fell considerably, with all but four of the emojis tested scoring a similarity of 0\%. Moreover, the four emojis that retained a similarity score above zero when the cosine similarity threshold was increased still saw decreases in similarity.

This further indicates the degree of difference in how emojis are semantically related to each other within the Anonymous and baseline datasets. There appears to be little relation in emoji-emoji definition between the datasets and, given that the cosine similarity thresholds used are relatively low, what relation there is, is likely based around neighbours that share fairly loose semantic relationships.

\subsection{Sentiment Analysis}

\begin{figure*}[!t]
\centering
\begin{subfigure}[b]{0.6\textwidth}
    \centering
    \includegraphics[width=\textwidth]{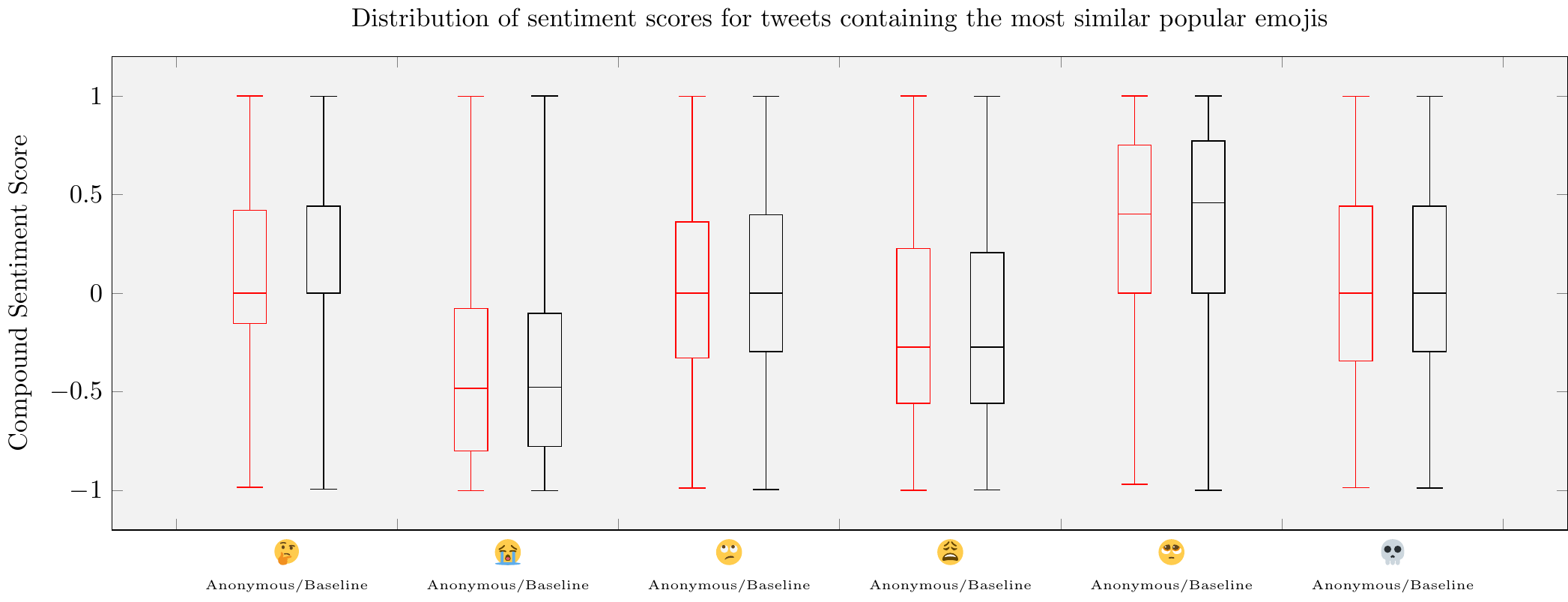}
    \caption{}
    \label{subfig:topEmojiSentiments}
\end{subfigure}
\quad
\begin{subfigure}[b]{0.6\textwidth}
     \centering
     \includegraphics[width=\textwidth]{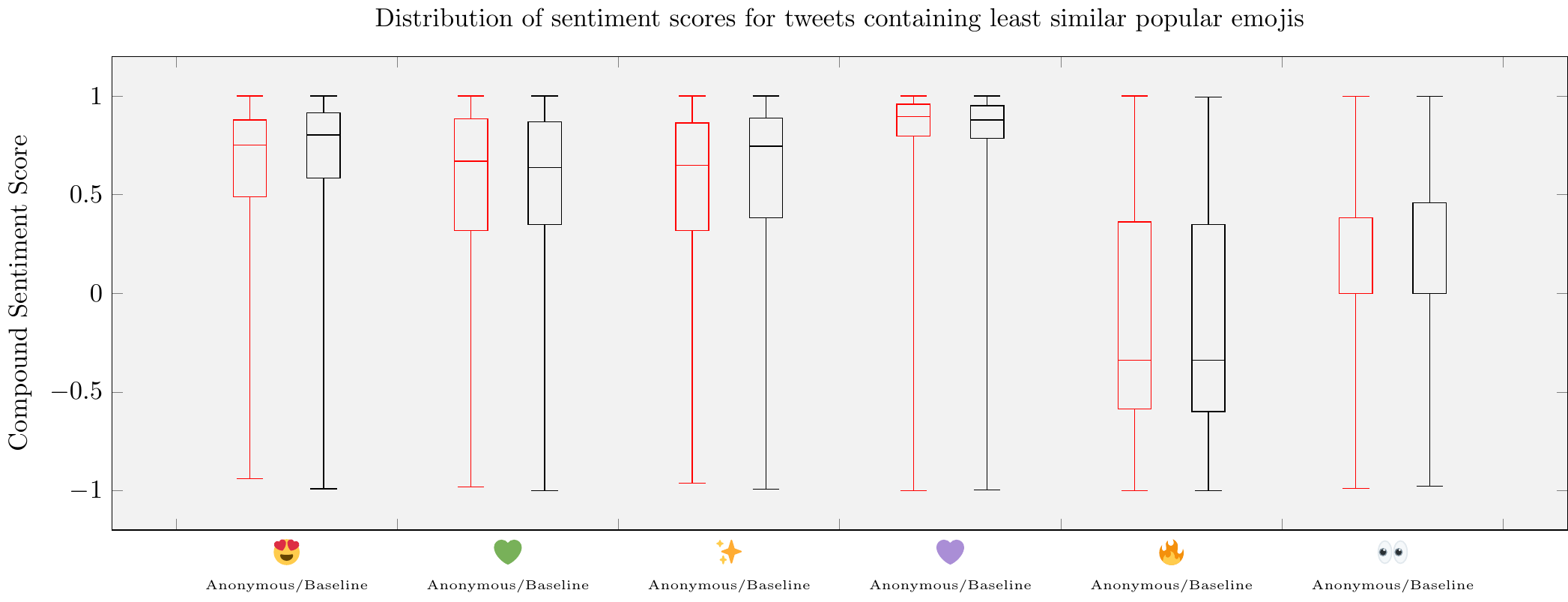}
     \caption{}
     \label{subfig:lowEmojiSentiments}
\end{subfigure}
\caption{Graphs showing the distribution of sentiments scores for Anonymous and baseline tweets containing certain emoji.}
\label{fig:emojiSentiments}
\end{figure*}

Although our findings so far point to differences in the manner in which emojis are used, one key factor of emoji usage is their role in strengthening emotional communication. Emojis provide an attempt at ubiquity that can, in theory, allow for the expression of emotions in a shared manner across a diverse range of groups~\cite{lu2016}. Given the key role of emojis in presenting emotion in a common manner, it is thus crucial that we examine the use of emojis in the context of the emotion being conveyed. 

We report Cohen's D for sentiments from tweets containing specific emojis drawn from Anonymous and non-Anonymous tweets in Table~\ref{table:EmojiSentimentCohenD} for the 25 most popular emojis from the two datasets. Scores of 0.2 or less are typically considered to constitute little to no effect, scores around 0.5 medium effect, and scores around 0.8 or greater large effect~\cite{cohen2013}.

\begin{table}[hb!]
\centering
\begin{tabular}{c | c | c} 
 \toprule
  \multicolumn{3}{c}{Emoji (Cohen's D)}\\
 \midrule
\thumbsUp~(0.20) & \thinkingFace~(0.09) & \greenHeart~(0.03)\\
\sparkles~(0.20) & \foldedHands~(0.08) & \smilingEyes~(0.02)\\
\blowingKiss~(0.18) & \pleadingFace~(0.07) & \hundredPoints~(0.02)\\
\twoHearts~(0.18) & \tearsJoy~(0.07) & \purpleHeart~(0.02)\\
\heartEyes~(0.16) & \rofl~(0.06) & \loudlyCrying~(0.01)\\
\eyes~(0.15) & \rollingEyes~(0.05) & \skull~(0.01)\\
\smilingFaceHearts~(0.16) & \redHeart~(0.05)~& \beamingEmoji~(0.00)\\
\winkingFace~(0.13) & \fire~(0.05)\\
\sunglasses~(0.10) & \wearyFace~(0.04)\\
\bottomrule
\end{tabular}
\caption{Using Cohen's D to compare sentiments of tweets containing a specific emoji from our datasets.}
\label{table:EmojiSentimentCohenD}
\end{table}

In Fig.~\ref{fig:emojiSentiments}, we also compare the distributions of sentiment for Anonymous and baseline tweets containing a given emoji. In Fig.~\ref{subfig:topEmojiSentiments}, we detail the results for the six emojis in Table~\ref{table:EmojiJaccardSim} that received the highest Jaccard similarity scores. In Fig.~\ref{subfig:lowEmojiSentiments}, we show the result for the six emojis that received the lowest Jaccard similarity scores. We focus on emojis with the highest and lowest Jaccard Index scores to offer insights into any potential relationships between emoji-emoji similarity and emoji sentiments. Sentiment scores are given on a scale of -1 to 1, with a score of -1 designating a highly negative tweet, a score of 0 a neutral tweet, and a score of 1 a positive tweet.

As we can see from Table~\ref{table:EmojiSentimentCohenD} and Fig.~\ref{fig:emojiSentiments}, there appears to be little difference in the typical sentiment of emojis between the two datasets. Interestingly, the tails of sentiment are large for all emojis in Fig.~\ref{fig:emojiSentiments}, as are the interquartile ranges (IQR) for most emoji. This indicates that whilst the spread of sentiment is very similar between our datasets, the sentiment being expressed in tweets containing these emojis is less predictable. This is perhaps surprising. Given emojis' role as a means of conveying emotion, one would perhaps expect each emoji to have a clearly defined emotional context in which it appears. Instead, bar a few exceptions such as the Sparkle (\sparkles) emoji, most emoji seem to have a fairly flexible sentiment context, with IQRs often crossing the boundary between negative and positive sentiments (e.g., the \rollingEyes~emoji and the \skull~emoji). This makes the apparent parallels in emoji sentiment between Anonymous and baseline datasets all the more surprising. Despite emojis in both datasets often taking on a range of sentiments, these ranges are very similar. This indicates commonality not just in terms of emoji as a singular means of expressing emotion, but commonality as a means of being able to express differing degrees of sentiment and even entirely different polarities of sentiment across these disparate groups.

What is additionally interesting is that this degree of similarity in sentiment seems to be unaffected by the Jaccard Index. In Fig.~\ref{subfig:lowEmojiSentiments} and Table~\ref{table:EmojiSentimentCohenD}, we see that these emojis, despite sharing few neighbours between datasets, still share similar distributions of sentiment.

Thus, whilst we discover that emoji usage by Anonymous accounts seems to share little similarity in emoji-emoji definitions when compared with non-Anonymous accounts, this does not necessitate a difference in their usage to convey emotion. Through this analysis of sentiment, we find that the popular emojis used in both datasets are typically used to express similar emotions, regardless of their nearest neighbour similarity. This lends additional credence to this notion of emoji as means of shared expression, indicating that whilst emojis between groups may share different relations to each other, they are still typically used to convey emotion in a similar manner. A notion that is further strengthened, given these emojis' typical abilities to operate in different sentiment contexts, in a consistent manner, across the two datasets.

\subsection{Context Analysis}

Currently, our study of Anonymous emoji usage has found somewhat contradictory results. To lend further insights into these findings, we examine the relationship between emojis and the text content of the tweets themselves. By examining this, we hope to identify the presence of any unique themes or topics that distinguishes emojis use in Anonymous and non-Anonymous tweets.

\begin{table*}[ht!]
\scriptsize
\centering
\begin{tabular}{c l | l} 
 \toprule
 \textbf{Emoji} & \textbf{Nearest Text neighbours} & \textbf{Context}\\
 \midrule
\thinkingFace & hmmm,hmmmm,hmmmmmm &
reflection, consideration, thinking\\
\midrule
\loudlyCrying & omgg,samee,stoppp,plss,wtff,mysticmessenger,bruhh,stopppp,sameee,wheezing &
hilarity, sadness, strong emotional reactions\\
\midrule
\rollingEyes & riiight,bich,smh,yannie & 
sarcasm, exasperation\\
\midrule
\wearyFace & uppp,sexc,fuckkkk,stopppp,uggh,ughh,daddyyyyy,ilyy,omgggg,daddyyy &
infatuaion, arousal\\
\midrule
\pleadingFace & cuteeee,awee,cuteee,pleasee,youu,uuu,awh,ilysm,youuuuu,muah &
cuteness, argument instigation/escalation, pleading, love\\
\midrule
\skull & lmfaooo,whattt,lmfaooooo,lmfaoooo,broooo,omgg,wheezing,bruh,stoppp,wtfff &
hilarity\\
\midrule
\heartEyes & sexc,daddyyy,zaddy,omgggg,kingggg,smexy,yassss,yass,daddyyyyy,daddyyyy &
arousal, infatuaion,\\
\midrule
\greenHeart & godblessusall,sista,fanks,frnd,daddyyyy & 
love, gratitude\\
\midrule
\sparkles & periodtt,gravestone,daddyyyyy,tingz,zaddy,daddyyy,sexc,luvs,periodttt & 
infatuaion, Trump death wish\\
\midrule
\purpleHeart & hoseok,loveislove,hobi,cuteeee,imwithaewwomen,jhope,taehyung,timkindness,tmkindness & 
love, K-Pop, LGBT support\\
\midrule
\fire & scprimary,scdebate,bernsquad,complementaryeducation & 
politics, Bernie Sanders\\
\midrule
\eyes & yessirrr, zillionbeers, summ, okayyy &
direct address\\
 \bottomrule
\end{tabular}
\caption{The most similar text tokens W2V neighbors to each emoji in Anonymous dataset tweets.}
\label{table:EmojiMostSimilarAnon}
\end{table*}

\begin{table*}[ht!]
\scriptsize
\centering
\begin{tabular}{c l | l} 
 \toprule
 \textbf{Emoji} & \textbf{Nearest Text neighbours} & \textbf{Context}\\
 \midrule
\thinkingFace & hmmm & 
reflection, consideration,\\
\midrule
\loudlyCrying & helppp,stoppppp,themmm,omggggggg,pleaseeeee,sameeee,whyyy,stoppp,helpp,ihy & 
hilarity, sadness, strong emotional reaction\\
\midrule
\rollingEyes & duhh,btchs,cuffed,misspell & 
sarcasm, exasperation\\
\midrule
\wearyFace & bitchhhh,whewww,badddd,lordddd,whewwww,okayyyyy,affff,ihy,niceeee & 
exasperatiion, weariness, arousal, love\\
\midrule
\pleadingFace & protecc,adorableeee,thankyouuuu,aaaaaaaa,aaaaaaaaaaaaaa,bhie,muchhh,muchhhh,awee & 
adoration, cuteness, gratitude, excitement\\
\midrule
\skull & bruhhhh,lmfaoooooo,lmfaooo,bitchhh,lmfaooooooo,lmaoooo,deadddd,lmao,wheezing & 
hilarity\\
\midrule
\heartEyes & loveeeeee,beautifull,fineeee,prettyyy,,prettyyyyyy & 
love, attraction\\
\midrule
\greenHeart & xx,thank,cutiepie,x,love,macha,thaank,mashaallah,brightwin,mashallah & 
love, gratitude\\
\midrule
\sparkles & No text items & N/A\\
\midrule
\purpleHeart & borahae,thankyoubts,happybirthdaytaehyung,happyy,pooo,happyjhopeday,cuuute,monie & 
K-Pop, love, \\
\midrule
\fire & ayyyyy,hotties, sizzling,up,mahn & 
sexual attraction, \\
\midrule
\eyes & hooooo,sheeeeesh,orrrr & 
direct address\\
 \bottomrule
\end{tabular}
\caption{The most similar text tokens W2V neighbors to each emoji in baseline dataset tweets.}
\label{table:EmojiMostSimilarRandom}
\end{table*}

In the interest of space, we focus on the six most and least similar emojis based on their emoji-emoji Jaccard Index. For each emoji, we identify the nearest text neighbours in each W2V model, using a cosine similarity threshold of 0.5 to ensure a reasonable degree of relevance. We then select the top ten text neighbours (where ten neighbours exist, else all text neighbours are presented) for each emoji, for each model, and present them in Tables~\ref{table:EmojiMostSimilarAnon} and \ref{table:EmojiMostSimilarRandom}. For each emoji in each dataset, we then manually examine tweets containing the emoji and each semantically linked keyword and label each emoji with its typical semantic and topic contexts.

From these results, we observe a large degree of consensus in the manner in which emojis are used by Anonymous and baseline accounts. For the majority of emojis, it seems the general context in which they are used is similar. Moreover, this similarity does not seem to depend heavily on the type of emojis being used. There also seems to be little relationship between the similarities in emoji-emoji definition noted in Section~\ref{secton:similarity} and the similarities in text context.

What is interesting is that this similarity in usage remains even though the topics focused on in the tweets vary considerably. In the Anonymous dataset, the \rollingEyes~emoji is often used to express exasperation at political events, such as \textit{``...Funny...in a Blue State... \thinkingFace\monocleEmoji\rollingEyes''} and \textit{``Told them they should advertise that they’re not going to sell to Trump supporters...smh \rollingEyes\smilingEyes''}. Whereas, in the baseline dataset the events are far more varied and less specific: \textit{``God's time is the best \rollingEyes\tearsJoy''}, and \textit{``Because she’s a BAD BITCH \rollingEyes~duhh''}. Despite the drastic array of topics focused on, with the Anonymous group unsurprisingly revealing a more focused set of typically political topics appropriate the to the hacktivist-based nature of the group, this seems to have little effect on how emojis are used.

Another surprising similarity, given its specificity in use, occurs with the Purple Heart (\purpleHeart) emoji. In the Anonymous accounts we note it is used in tandem with references to various K-Pop (Korean popular music) figures. This can be seen in the related terms in Table~\ref{table:EmojiMostSimilarAnon}, with references to notable K-Pop figures such as J-Hope, and Kim Tae-hyung. Whilst initially a link between K-Pop and Anonymous may seem strange, this is not necessarily surprising as Twitter accounts affiliated with K-Pop fandom have been noted for declaring their support for Anonymous during the summer of 2020 in support of Black Lives Matter, and the protests over the killing of George Floyd by a Minneapolis Police Department officer~\cite{Independant2020}.

What is curious, however, is this same link to K-Pop appears in the use of the Purple Heart emoji within the baseline dataset. This is a surprising similarity, given the specificity of the usage to this specific genre of music. Furthermore, additional research revealed a likely link between the purple heart and a phrase associated with the popular K-Pop group BTS (of which Kim Tae-hyung is the lead singer): ``I purple you''~\cite{newsweek2019}. This is further evidenced by tweets in our baseline dataset, including \textit{``I purple u too \purpleHeart''}. We thus see evidence of the surprising development of a new usage of the emoji, in part linked to accounts declaring an affinity to Anonymous, indicating that members of this fandom have aligned themselves with the group. In turn, bringing with them this evolution in emoji usage.

Within these similarly used emoji however, there are still some interesting differences. With the Heart-Eyes (\heartEyes) emoji, although the general sentiments is focused on love and attraction, they are often manifested quite differently in the two datasets. In the baseline dataset, the emoji is generally used to express general adoration and/or love, e.g., \textit{``I loveeeee \heartEyes~my favorite one''}, and \textit{``u r so prettyyy \heartEyes''}. In the Anonymous dataset, however, this expression of love is often done in a more unconventional manner. Instead, this emoji appears to be used by accounts to express infatuation, and even sexual attraction, towards some of the more prominent Anonymous accounts in the network. For instance, we see tweets such as \textit{``@YourAnonCentral Yesssssssss you are our daddyyyysssss \twoHearts\heartEyes\heartEyes''} and \textit{``@YourAnonNews you can it daddy \heartEyes''}. The infatuation and borderline fetishism accompanying this emoji, whilst in essence similar to its use in the borderline dataset, presents a very extreme manifestation of this usage relative to the baseline dataset. 

These findings can also be seen in the Weary (\wearyFace) emoji. Again, whilst there are clear similarities in uses between Anonymous and baseline accounts: using the emoji to express attraction, the Anonymous usage again focuses primarily around infatuation with central Anonymous accounts, with tweets such as \textit{``@YourAnonNews...sexy...Daddy Anon \wearyFace\crazyEyes\droolingEmoji''} and \textit{``@YourAnonNews daddy anon \wearyFace\smilingFaceHearts''}. 

These findings lend interesting context to the results in \citeauthor{Jones2020}'s study~(\citeyear{Jones2020}) of the group on Twitter, which found evidence of the group's centralisation around a small number of accounts. It was suggested in their work that this was contrary to the group's aims of having a decentralised, leaderless network -- a not unreasonable claim given the group's stated philosophy in the past in which they rejected notions of hierarchy~\cite{Uitermark2017}. However, from this it appears that at least a reasonable contingent of Anonymous accounts (given the key terms associated with the \heartEyes~emoji, and the prevalence of the emoji as noted in Table~\ref{table:emojiFrequencies}) actually seem to be content with supporting, to an extreme extent, these central Anonymous accounts. A finding which indicates some sense of evolution in the group's dynamic and central philosophies.

In turn, we find that the context in which these popular emojis are actually surprisingly similar between Anonymous and non-Anonymous users. However, we also find evidence of interesting specified usage in some of these emojis, where Anonymous users leverage the `typical' meaning of certain emojis, commonly associated with expressing affection, and exaggerate and focus them on declaring some sense of infatuation with more prominent members of the group. This finding calls into question the extent to which members of the group reject notions of leaderlessness.

\section{Conclusion}

In summary, we identify a relationship in emoji usage between Anonymous Twitter accounts and `typical' Twitter accounts, which strengthens the notion of emojis as a `ubiquitous' language~\cite{lu2016}. 

Although, through our study of the emoji-emoji semantic relations in the two groups we find good evidence that the semantic relationships between emojis differ, further analysis indicates that this does not necessitate difference in practical usage. This highlights to researchers the potential limitations in relying solely on this form of analysis, demonstrating that additional metrics are needed to gain an appreciation of a group's overall use of emojis.

In turn, we note in our study of sentiment that emojis are used in very similar ways in Anonymous and non-Anonymous tweets as a means of expressing emotion. A finding that is particularly insightful given the range of sentiments that many emojis appear to be used to convey, and the similarities in these ranges between Anonymous and non-Anonymous accounts. Moreover, our study of the text items that share the closest semantic links to popular emojis also reveal similarities in emoji usage. Despite the Anonymous accounts sharing an alignment in interests that differs from `typical' Twitter users, it seems that emoji usage has maintained a recognisable sense of commonality within the tweets of Anonymous-affiliated accounts.

Despite this, we identify evidence of nuances in the Anonymous usage of emojis that provide insights into the group's activity. We observe that certain emojis have received unusual, narrower uses in Anonymous tweets, particularly as a means of expressing infatuation with the more prominent Anonymous Twitter accounts. Not only does this specification indicate a shift in usage, it also indicates a shift in the group's ethos. Whilst Anonymous typically has presented itself as anti-hierarchy, this emoji usage indicates that at least among some users this feeling has shifted to something more resembling infatuation with these centralised affiliates. This finding emphasises the ephemeral nature of the group, and demonstrates the insights that can be gleaned via the study of emoji usage.

\subsection{Limitations and Future Work}

This study does come with limitations that should be acknowledged. Firstly, when interpreting these results we must consider that in both datasets there is a degree of difference in the tweet output of different accounts, an imbalance which may lead to bias towards the more active accounts in the dataset. This is a difficult problem as there is little that can be done to account for the tweeting behaviours of users that tweet infrequently. However, it is a factor that must be considered when attempting to generalise from our results. Additionally, whilst the use of computational models is useful as a means of offering a broad understanding of emoji usage in our datasets, this approach does risk losing the nuance present in a given group's use of emojis.

In future, therefore, a project that aims at conducting a large-scale qualitative study of emoji usage in Anonymous tweets would be useful. Supplementing our method with a qualitative approach would help mitigate the weaknesses in using large-scale summative models, providing additional detail to the broader findings presented here. Moreover, additional analysis of emoji usage by other hacktivist groups would be of interest. Whilst Anonymous is interesting due to its notoriety and public image, examining whether similar patterns of emoji usage are present in other hacktivist groups would be of great interest. This could also be expanded to other online groups, such as QAnon or Black Lives Matter.


\fontsize{9.5pt}{10.5pt} 
\selectfont
\bibliography{FULL-JonesK} 

\end{document}